\documentclass{article}

\PassOptionsToPackage{numbers, compress}{natbib}

\usepackage[preprint]{neurips_2020}

\usepackage[utf8]{inputenc} 
\usepackage[T1]{fontenc}    
\usepackage{hyperref}       
\usepackage{url}            
\usepackage{booktabs}       
\usepackage{amsfonts}       
\usepackage{nicefrac}       
\usepackage{microtype}      

\usepackage{amsmath}
\usepackage{graphicx}
\usepackage{rotating}
\usepackage{xcolor}

\title{Improving Style-Content Disentanglement in Image-to-Image Translation}

\author{
Aviv Gabbay ~~~~~~~ Yedid Hoshen \\
School of Computer Science and Engineering \\
The Hebrew University of Jerusalem, Israel \\
}

\begin{document}

\maketitle

\begin{abstract}
Unsupervised image-to-image translation methods have achieved tremendous success in recent years. However, it can be easily observed that their models contain significant entanglement which often hurts the translation performance. In this work, we propose a principled approach for improving style-content disentanglement in image-to-image translation. By considering the information flow into each of the representations, we introduce an additional loss term which serves as a content-bottleneck. We show that the results of our method are significantly more disentangled than those produced by current methods, while further improving the visual quality and translation diversity.
\end{abstract}

~~~~~~~~~~~~~~~~~~Project webpage: \textcolor{blue}{http://www.vision.huji.ac.il/style-content-disentanglement}

\section{Introduction}
Image translation is the task of mapping images between different domains, i.e. given an input image in a source domain (e.g. dogs), we aim to generate an analogous image in a target domain (e.g. cats). Although this task is generally poorly specified, it is often made possible under the assumption that images in different domains share similar \textit{content} (e.g. head pose) which can be transferred over during translation. In cases where pairwise correspondences between domains are available, general-purpose conditional adversarial networks such as pix2pix \cite{isola2017image, wang2018high} achieve remarkable results and scale up to extremely high resolutions. Much current research in image translation deals with the more challenging unsupervised setting in which no correspondences are given. Early attempts \cite{zhu2017unpaired} to solve this problem find a single image in the target domain for every input image in the source domain. While this one-to-one mapping can be satisfactory in some cases, most image domains are multi-modal in nature i.e. there are several possible mappings for every input (e.g. an image of a cat can be translated to images of different dog breeds). As a result, uni-modal formulations fail to capture the underlying distribution of images in different domains. Notable methods such as MUNIT \cite{huang2018multimodal} and StarGAN-v2 \cite{choi2019stargan} tackle the multi-modal translation task and present high quality and diverse domain mappings along with reference-guided image synthesis in which the specific target \textit{style} is borrowed from a reference image in the target domain.

Another closely related line of work studies the problem of learning disentangled representations in the class-supervised setting \cite{gabbay2019demystifying, denton2017unsupervised, bouchacourt2018multi}. In this task, the goal is to learn disentangled representations for each class (domain) in the dataset and a residual content representation for each image. For example, LORD \cite{gabbay2019demystifying} utilizes latent optimization for learning a class representation that is shared exactly between all images of the same class and an additional regularized representation which captures the image-specific content and can be applied to images from different classes. It is shown that non-adversarial bottlenecks provide better disentanglement than methods that rely on domain confusion losses. In this work, we draw inspiration from these principles and carefully analyse the information between the \textit{domain}, \textit{content}, and \textit{style} representations. We introduce an additional loss term which strongly encourages content-style disentanglement. We then show that state-of-the-art architectures for image translation can greatly benefit from these disentanglement principles to achieve higher translation quality and greater output diversity.

\section{Related Work}
\paragraph{Image Translation} \citet{isola2017image} propose pix2pix, a conditional generative adversarial network, as a general purpose model for solving different image-to-image tasks in the supervised setting. \citet{wang2018high} extend this framework to generating high resolution images. Initial progress in the unsupervised setting has been made by CycleGAN \cite{zhu2017unpaired} which introduces a cycle consistency loss to guarantee that the translated image properly preserves the domain-invariant characteristics (e.g. pose) of the source image. As a consequence of this approach, the model learns a deterministic one-to-one mapping, thus can not capture the multi-modal nature of the image distribution.  MUNIT \cite{huang2018multimodal} recognizes this limitation and extends the framework to learn multi-modal mappings. Despite its capability of generating diverse and realistic translation outputs, this method trains separate encoder-decoder models for each domain and therefore can not easily scale up to multiple domains. StarGAN \cite{choi2018stargan,choi2019stargan} addresses this issue and presents a unified model for multi-domain translation. FUNIT \cite{liu2019few} attempts to generalize to images from unseen domains using a few reference images from a target domain, but it requires fine-grained class labels during training and can not model intra-class unspecified variations.

\paragraph{Class-Supervised Disentanglement} Learning disentangled representations from a set of observations is a fundamental problem in machine learning. The most related setting to our work is the class-supervised setting in which there exists a class label for every image. The goal is generally to anchor the semantics of all the images within each class into a separate class representation while modeling all the remaining image-specific properties by a content representation per image. Several methods encourage disentanglement by adversarial constraints \cite{denton2017unsupervised, szabo2017challenges, mathieu2016disentangling} while other rely on cycle consistency \cite{harsh2018disentangling} or group accumulation \cite{bouchacourt2018multi}. LORD \cite{gabbay2019demystifying} takes a non-adversarial approach and trains a generative model while directly optimizing over class and content codes. Although most of the works in this area demonstrate domain translation results, they do not deal with the multi-modal unsupervised setting. Moreover, their primary objective is to achieve disentanglement at the representation level and put less effort on tuning architectures for high-quality image translation. 

\section{An Analysis of Disentanglement in Image-Translation}
\label{sec:analysis}

Let us model the formation of an image $x_i$ as a function of domain $y_i$, style $s_i$ and content $c_i$.

\begin{equation}
    \label{eq:formation}
    x_i = G(y_i, s_i, c_i)~~~~ x_i\in\mathcal{X}~~~y_i\in\mathcal{Y}~~~s_i\in\mathcal{S}~~~c_i\in\mathcal{C}
\end{equation}

Each image $x_i$ belongs to a single domain $y_i$ (e.g. "cat" or "dog"). The content $c_i$ describes the information that is invariant across domains (e.g. head pose). This information should be preserved if the domain label of the image is changed. The style $s_i$ describes the residual properties that are not preserved across domains and are not shared across all images in the same domain i.e. intra-domain variations. As an illustrative example, let us assume that we are provided with a set of images each classified into either "cat" or "dog". The domain specifies the characteristics which are shared across all cats or dogs. The content describes the information which is invariant to the species e.g. the pose of the animal. The style captures the information not shared across species or shared by all member of the same species e.g. the breed of the animal, its color or the texture of its fur. 

Current image translation models encourage the disentanglement of these three factors of variation by introducing several different constraints. Let us briefly review the most common techniques.

The most common approach for learning a domain-invariant content representation is by using a domain confusion objective. MUNIT \cite{huang2018multimodal} trains a discriminator for each of the domains which aims at distinguishing whether the output image is from the specific domain or not. StarGAN-v2 \cite{choi2019stargan} scales this approach to multiple domains and trains a conditional discriminator which learns to identify whether the output image is from the given target domain.

Alternatively, another popular approach is by utilizing a conditional discriminator at the representation level of the content code which attempts to predict the domain label given the content code \cite{benaim2019domain, denton2017unsupervised}.
There are two issues with relying on adversarial constraints for disentanglement: i) GAN training is unstable and sensitive to hyper-parameters due to the challenging saddle point optimization problem. It is shown \cite{gabbay2019demystifying} that GAN discriminators often do not in fact remove all domain-specific information from the content representation. ii) The conditional discriminators only ensure disentanglement between the content and the \textit{domain} but not necessarily between the content and the \textit{style}. As we do not have supervision on the style, it is not obvious how to train a style conditional discriminator. The content codes may therefore contain a significant amount of style information. 

In the other direction, in order constrain the capacity of the style representation and avoid leakage of content information, current methods rely on locality-preserving architectures that bias translations towards local changes. The style is typically injected as scale and shift parameters into Adaptive Instance Normalization (AdaIN) \cite{huang2017arbitrary, karras2019style} layers at different levels of the generator architecture. As this operates in a global manner per channel, it effectively preserves the spatial structure of the content image. Although this makes optimization easier, it inevitably limits the diversity of generated images which is typically constrained to low-level variations.

In this work, we focus on improving content-style disentanglement in image translation models, by introducing a well-motivated bottleneck on the content representation. We show that integrating this term with state-of-the-art architectures greatly improves the quality and diversity of translation results.


\section{Improved Content-Style Disentanglement with a Content Bottleneck}
\label{sec:method}

We provide a principled approach for improving content-style disentanglement in image translation. 

\subsection{Image Translation Framework}
Our architecture is strongly influenced by those of state-of-the-art methods such as StarGAN-v2 \cite{choi2019stargan} and MUNIT \cite{huang2018multimodal}. 

Given an input image $x \in \mathcal{X}$ in domain $y \in \mathcal{Y}$, we train a content encoder $E_c: \mathcal{X} \xrightarrow{} \mathcal{C}$ and a domain-conditional style encoder $E_s: \mathcal{X} \times \mathcal{Y} \xrightarrow{} \mathcal{S}$ to obtain a content code $c \in \mathcal{C}$ and a style code $s \in \mathcal{S}$, respectively:

\begin{equation}
\label{eq:encoders}
    c = E_c(x) ~~~~~ s = E_s(x, y)
\end{equation}

During training, we sample a random image $x$ from a source domain $y$ and optimize the generator network $G$ to translate $x$ into an image in a random target domain $\hat{y}$ with the style $\hat{s}$ of another randomly sampled image $\hat{x}$:

\begin{equation}
\label{eq:translation}
    x_{y \xrightarrow{} \hat{y}} = G(\hat{y}, \hat{s}, c) ~~~~~ \hat{s} = E_s(\hat{x}, \hat{y})
\end{equation}

In order to encourage $G$ to generate valid images from a target domain, we train a conditional discriminator $D$ and employ an adversarial loss:

\begin{equation}
\label{eq:adv}
    \mathcal{L}_{adv} = \mathbb{E}_{x, y}[\log D(x, y)] + \mathbb{E}_{x, \hat{x}, \hat{y}}[\log (1 - D(G(\hat{y}, \hat{s}, c), \hat{y}))]
\end{equation}


To enforce $G$ to preserve the content of the input image $x$, we apply a cycle reconstruction loss:

\begin{equation}
    \label{eq:recon}
    \mathcal{L}_{rec} = \mathbb{E}_{x, y, \hat{x}, \hat{y}} \| G(y, s, E_c(x_{y \xrightarrow{} \hat{y}})) - x\|_1
\end{equation}

\subsection{Content Bottleneck} 
In Sec.~\ref{sec:analysis} we argued that current methods do not explicitly ensure the disentanglement of the content from style. To this end, we propose a simple but principled and subtle new term to encourage disentanglement. Specifically, we regularize the content by turning the content encoder and the generator into a variational auto-encoder by introducing a KL-divergence loss between the content code and a prior Gaussian distribution. Although typically in VAE, the encoder learns both the mean and log-variance of the random variable, we follow LORD \cite{gabbay2019demystifying} in using noise with a constant (unlearned) variance. This prevents partial posterior collapses, and prevents information leakage through the content code. The additional content-bottleneck (cb) term is therefore: 


\begin{equation}
    \label{eq:kl}
    \mathcal{L}_{cb} = D_{KL}(\mathcal{N}(E_c(x), \sigma^2 I)~||~\mathcal{N}(0, \sigma^2 I))
\end{equation}

The content code is changed accordingly during the feed-forward step:

\begin{equation}
\label{eq:encoder_with_noise}
    c = E_c(x) + z ~~~~ z \sim \mathcal{N}(0, \sigma^2 I)
\end{equation}

Our entire objective can be summarized as:

\begin{equation}
    \label{eq:full}
    \min_{E_c, E_s, G} ~ \max_{D} ~~ \mathcal{L}_{rec} + \lambda_{adv} \mathcal{L}_{adv} + \lambda_{cb} \mathcal{L}_{cb}
\end{equation}

\subsection{Implementation Details}
In order to emphasize the contribution of our proposed loss term for the task of style-content disentanglement, and the simplicity of integrating it into existing state-of-the-art image translation models, we base our model on exact same architectures of all the neural networks proposed in StarGAN-v2 \cite{choi2019stargan}. We will briefly describe the main components for completeness.

\textbf{Content Encoder} Our content encoder takes as input an image $x$ and outputs a content code $c$. As the content is regularized within a VAE in our framework, we find that fixing the variance of the estimated distribution improves stability and avoids leakage of other information into the content.

\textbf{Style Encoder and Mapping Network} The style encoder takes two inputs: an image $x$ and a corresponding domain label $y$ and outputs a style code $s$. In order to enable sampling random styles we include a mapping network that translates a latent code $z$ sampled from a prior Gaussian distribution to a style code $s$ through a series of fully-connected layers. Note that the last layers are trained in a domain-specific fashion to improve performance.

\textbf{Generator} Our generator takes two inputs: a content code $c$ and a style code $s$. Note that since the style is already modulated with the domain $y$ there is no need to provide the generator with $y$ directly. Similarly to other state-of-the-art domain translation methods, the style is injected as scale and shift parameters into Adaptive Instance Normalization (AdaIN) \cite{huang2017arbitrary, karras2019style} layers at different levels of the generator architecture.

\textbf{Discriminator} The discriminator is domain-conditonal and therefore takes as input an image x and a domain label $y$ and determines whether the image is a real image of domain $y$ or fake image generated by the generator.

We set $\sigma = 1$ and $\lambda_{adv} = 1$, $\lambda_{cb} = 0.0001$ in all our experiments.

\section{Experiments}
\subsection{Baselines}
Our method is evaluated against StarGAN-v2 \cite{choi2019stargan} which is a state-of-the-art domain translation method and against the established MUNIT \cite{huang2018multimodal} framework. While StarGAN-v2 supports multiple domains similarly to our method, MUNIT is trained multiple times for every possible pair of domains in the following experiments.

\subsection{Datasets}
\paragraph{AFHQ} We first assess the performance of all methods on the recently proposed Animal-Faces-HQ dataset \cite{choi2019stargan}, consisting of $15,000$ high quality images. The images are categorized into three domains: cat, dog and wildlife. We follow the protocol used in \cite{choi2019stargan} and use 500 images from each domain as a test set and the rest as a training set.
\paragraph{CUB} To further compare the disentanglement performance of our method and StarGAN-v2 on multi-domain translation with tens of domains which exhibit a considerable amount of style variation, we compare both methods on CUB-200-2011 \cite{wah2011caltech}, a dataset of 200 bird species with 6,000 images. In order to increase the variance within each domain we aggregate similar bird species by their descriptions (e.g. Gull, Woodpecker) and form CUB-47 variant in which the birds are separated into only $47$ coarse domains, which is a much more challenging benchmark for multi-modal domain translation methods.

\paragraph{Edges2Shoes} \citep{yu2014fine}
A collection of 50,000 images separated into two domains: shoe images and their edges. Note that we do not make use of the pairwise correspondences provided in this dataset.

We conduct all the experiments at 128x128 resolution.

\subsection{Evaluation Protocol}
In order to measure the diversity of translation results, we follow \cite{choi2019stargan} and translate all images in the test set to each of the other domains multiple times using two strategies: i) \textit{Reference-guided} translation i.e. borrowing style codes from random reference images in the target domain. ii) \textit{Sampling}-based translation i.e. generating outputs by sampling different random style codes. We then measure the perceptual pairwise distances using LPIPS \cite{zhang2018unreasonable} between all translations of the same input image. Higher average distances indicate greater diversity in image translation. To assess the improvement in style-content disentanglement, we compute FID \cite{heusel2017fid} which measures the discrepancy between the distribution of images in each target domain and the corresponding translations generated by the models. A lower FID score indicates that the translations are more reliable and better fit to the target domain. 
It should be noted that none of the used metrics is well-suitable for measuring style-content disentanglement. LPIPS merely measures the amount of variation due to style, but not its fidelity to the original style or its disentanglement from the content. FID measures the similarity between the generated and true images in a particular domain, but it does not measure the similarity to a particular style. FID is therefore more effective for class-content disentanglement than multi-modal setting that contain style. Developing competent metrics for the content-style disentanglement setting is an important and unsolved task.

\subsection{Results}
\begin{table}[t]
  \centering
  \caption{Multi-modal domain translation results between the three domains of AFHQ.}
  \label{tab:afhq}
    \begin{tabular}{lcccc}
    \toprule
    & \multicolumn{2}{c}{Reference-guided} & \multicolumn{2}{c}{Sampling} \\
    \midrule
    
    & LPIPS $\uparrow$ & FID $\downarrow$ & LPIPS $\uparrow$ & FID $\downarrow$ \\
    \midrule
    MUNIT \cite{huang2018multimodal} & 0.302 & 139.732 & 0.426 & 32.208 \\
    StarGAN-v2 \cite{choi2019stargan} & 0.328 & 27.842 & 0.357 & 28.751 \\
    Ours & \textbf{0.394} & \textbf{23.927} & \textbf{0.427} & \textbf{18.767} \\
	\bottomrule
    \end{tabular}
\end{table}

\begin{table}[t]
  \centering
  \caption{Multi-modal domain translation results between 47 domains of CUB-47.}
  \label{tab:cub}
    \begin{tabular}{lcccc}
    \toprule
    & \multicolumn{2}{c}{Reference-guided} & \multicolumn{2}{c}{Sampling} \\
    \midrule
    
    & LPIPS $\uparrow$ & FID $\downarrow$ & LPIPS $\uparrow$ & FID $\downarrow$ \\
    \midrule
    StarGAN-v2 \cite{choi2019stargan} & 0.231 & 76.391 & 0.182 & \textbf{80.196} \\
    Ours & \textbf{0.315} & \textbf{71.738} & \textbf{0.239} & 93.688 \\
	\bottomrule
    \end{tabular}
\end{table}

\begin{figure}
\begin{center}

\begin{tabular}{@{\hskip0pt}c@{\hskip0pt}c@{\hskip0pt}c@{\hskip0pt}c@{\hskip0pt}c@{\hskip0pt}c}
& ~~~~~~~~~~~~~\textbf{Style} ~~~~~~ \begin{turn}{90} ~~~~~~~ \textbf{Content}\end{turn} ~ & \includegraphics[width=0.19\linewidth]{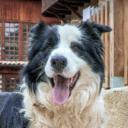} & \includegraphics[width=0.19\linewidth]{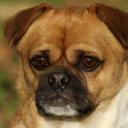} &
\includegraphics[width=0.19\linewidth]{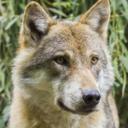} & \includegraphics[width=0.19\linewidth]{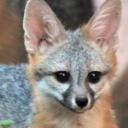} \\

\begin{turn}{90} ~~~~ StarGAN-v2\end{turn} & 
\includegraphics[width=0.19\linewidth]{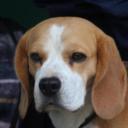} & \includegraphics[width=0.19\linewidth]{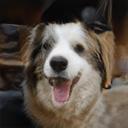} & \includegraphics[width=0.19\linewidth]{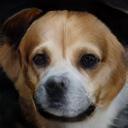} & \includegraphics[width=0.19\linewidth]{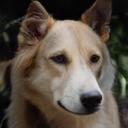} &
\includegraphics[width=0.19\linewidth]{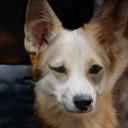} \\
\begin{turn}{90} ~~~~~~~~~ \textbf{Ours}\end{turn} & 
 &
 \includegraphics[width=0.19\linewidth]{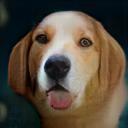} & \includegraphics[width=0.19\linewidth]{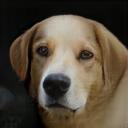} & \includegraphics[width=0.19\linewidth]{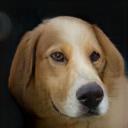} &
\includegraphics[width=0.19\linewidth]{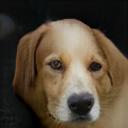} \\

\begin{turn}{90} ~~~~ StarGAN-v2\end{turn} & 
\includegraphics[width=0.19\linewidth]{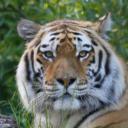} & \includegraphics[width=0.19\linewidth]{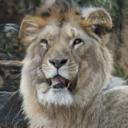} & \includegraphics[width=0.19\linewidth]{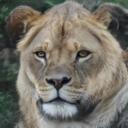} & \includegraphics[width=0.19\linewidth]{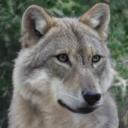} &
\includegraphics[width=0.19\linewidth]{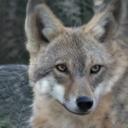} \\
\begin{turn}{90} ~~~~~~~~~ \textbf{Ours}\end{turn} & 
 &
 \includegraphics[width=0.19\linewidth]{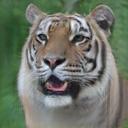} & \includegraphics[width=0.19\linewidth]{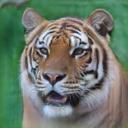} & \includegraphics[width=0.19\linewidth]{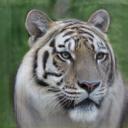} &
\includegraphics[width=0.19\linewidth]{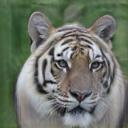} \\

\begin{turn}{90} ~~~~ StarGAN-v2\end{turn} & 
\includegraphics[width=0.19\linewidth]{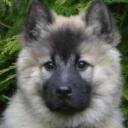} & \includegraphics[width=0.19\linewidth]{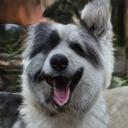} & \includegraphics[width=0.19\linewidth]{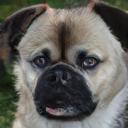} & \includegraphics[width=0.19\linewidth]{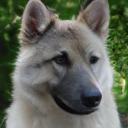} &
\includegraphics[width=0.19\linewidth]{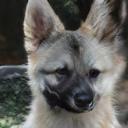} \\
\begin{turn}{90} ~~~~~~~~~ \textbf{Ours}\end{turn} & 
 &
 \includegraphics[width=0.19\linewidth]{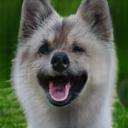} & \includegraphics[width=0.19\linewidth]{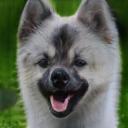} & \includegraphics[width=0.19\linewidth]{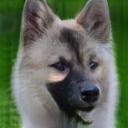} &
\includegraphics[width=0.19\linewidth]{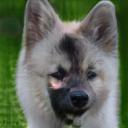} \\

\end{tabular}
\end{center}
\caption{Comparison between our method and StarGAN-v2 on AFHQ. StarGAN-v2 leaks a significant amount of details of the content image and generates unreliable and inconsistent translations. Our method produces much more disentangled results and captures the target style faithfully.}
\label{fig:afhq_ref}
\end{figure}

We present a quantitative evaluation of our method and state-of-the-art baseline methods MUNIT and StarGAN-v2 on the AFHQ dataset in Tab.~\ref{tab:afhq}. It can be seen that our method achieves both higher translation diversity (LPIPS) and better visual quality (FID) than both baselines in the reference-guided setting as well as when sampling random style codes. It should be noted that as observed in \cite{choi2019stargan}, in the case of reference-guided translation, MUNIT experiences mode-collapse and fails to generate diverse images according to provided reference images. As can be seen in the qualitative examples in Fig.~\ref{fig:afhq_ref}, StarGAN-v2 leaks a significant amount of details from the input image which results in unreliable translation between the different domains.

Furthermore, it is mostly capable of changing low-level features as color and texture. Our method generates much more disentangled translations and captures the exact style of the reference image.

\begin{figure}[t]
\begin{center}

\begin{tabular}{@{\hskip0pt}c@{\hskip0pt}c@{\hskip0pt}c@{\hskip0pt}c@{\hskip0pt}c@{\hskip0pt}c}
& ~~~~~~~~~~~~~\textbf{Style} ~~~~~~ \begin{turn}{90} ~~~ \textbf{Content}\end{turn} ~ & \includegraphics[width=0.13\linewidth]{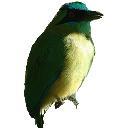} & \includegraphics[width=0.13\linewidth]{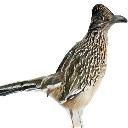} &
\includegraphics[width=0.13\linewidth]{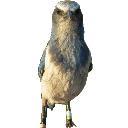} & \includegraphics[width=0.13\linewidth]{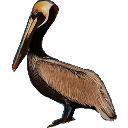} \\

\begin{turn}{90} ~ StarGAN-v2\end{turn} & 
\includegraphics[width=0.13\linewidth]{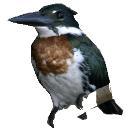} & \includegraphics[width=0.13\linewidth]{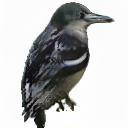} & \includegraphics[width=0.13\linewidth]{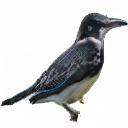} & \includegraphics[width=0.13\linewidth]{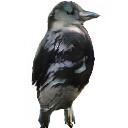} &
\includegraphics[width=0.13\linewidth]{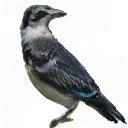} \\
\begin{turn}{90} ~~~~~~ \textbf{Ours}\end{turn} & 
 &
 \includegraphics[width=0.13\linewidth]{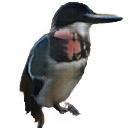} & \includegraphics[width=0.13\linewidth]{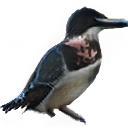} & \includegraphics[width=0.13\linewidth]{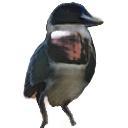} &
\includegraphics[width=0.13\linewidth]{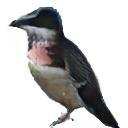} \\

\begin{turn}{90} ~ StarGAN-v2\end{turn} & 
\includegraphics[width=0.13\linewidth]{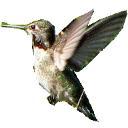} & \includegraphics[width=0.13\linewidth]{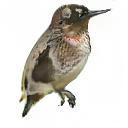} & \includegraphics[width=0.13\linewidth]{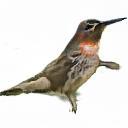} & \includegraphics[width=0.13\linewidth]{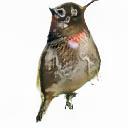} &
\includegraphics[width=0.13\linewidth]{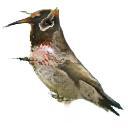} \\
\begin{turn}{90} ~~~~~~ \textbf{Ours}\end{turn} & 
 &
 \includegraphics[width=0.13\linewidth]{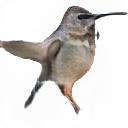} & \includegraphics[width=0.13\linewidth]{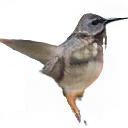} & \includegraphics[width=0.13\linewidth]{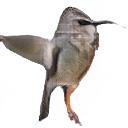} &
\includegraphics[width=0.13\linewidth]{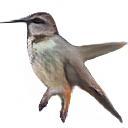} \\

\end{tabular}
\end{center}
\caption{Comparison between our method and StarGAN-v2 on CUB-47. Our method better captures the fine-grained details of the target style. For example, it is able to transfer-over the breast color and the bill shape.}
\label{fig:cub_ref}
\vspace{-1em}
\end{figure}

In order to assess the performance of the methods in the multi-domain translation setting, we follow the same protocol on the CUB-47 dataset. In this dataset, the images are classified into 47 different bird species, while the images in each domain exhibit variations in fine-grained details. Results of this experiments are presented in Tab.~\ref{tab:cub} along with visual examples in Fig.~\ref{fig:cub_ref}. In this experiment we compare our method only to StarGAN-v2, as the MUNIT architecture supports only a pair of domains and can not scale to this amount of domains. It can be seen that we achieve higher diversity and better visual quality than StarGAN-v2 in the reference-guided setting although exhibit some degradation in quality in the sampling case. Moreover, out method is capable of translating to a specific fine-grained style of the bird, as it correctly applies the breast color and the bill shape.

\begin{figure}[t]
\begin{center}

\begin{tabular}{@{\hskip0pt}c@{\hskip0pt}c@{\hskip0pt}c@{\hskip0pt}c@{\hskip0pt}c@{\hskip0pt}c}
& ~~~~~~~~~~~~~\textbf{Content} ~~~~~~ \begin{turn}{90} ~~~~ \textbf{Style}\end{turn} ~ & \includegraphics[width=0.13\linewidth]{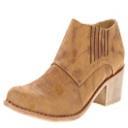} &
\includegraphics[width=0.13\linewidth]{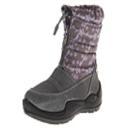} & \includegraphics[width=0.13\linewidth]{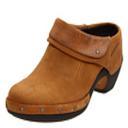} &
\includegraphics[width=0.13\linewidth]{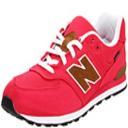} \\

\begin{turn}{90} StarGAN-v2\end{turn} & 
\includegraphics[width=0.13\linewidth]{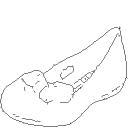} & \includegraphics[width=0.13\linewidth]{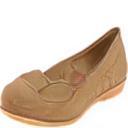} & \includegraphics[width=0.13\linewidth]{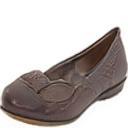} & \includegraphics[width=0.13\linewidth]{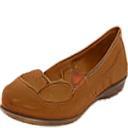} &
\includegraphics[width=0.13\linewidth]{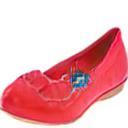} \\
\begin{turn}{90} ~~~~ \textbf{Ours}\end{turn} & 
 &
 \includegraphics[width=0.13\linewidth]{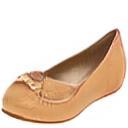} & \includegraphics[width=0.13\linewidth]{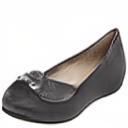} & \includegraphics[width=0.13\linewidth]{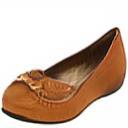} &
\includegraphics[width=0.13\linewidth]{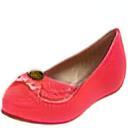} \\

\begin{turn}{90} StarGAN-v2\end{turn} & 
\includegraphics[width=0.13\linewidth]{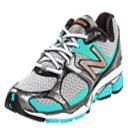} & \includegraphics[width=0.13\linewidth]{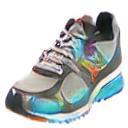} & \includegraphics[width=0.13\linewidth]{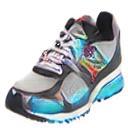} & \includegraphics[width=0.13\linewidth]{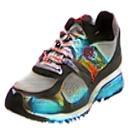} &
\includegraphics[width=0.13\linewidth]{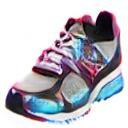} \\
\begin{turn}{90} ~~~~ \textbf{Ours}\end{turn} & 
 &
 \includegraphics[width=0.13\linewidth]{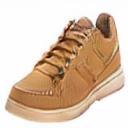} & \includegraphics[width=0.13\linewidth]{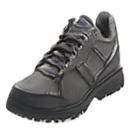} & \includegraphics[width=0.13\linewidth]{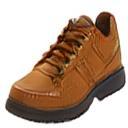} &
\includegraphics[width=0.13\linewidth]{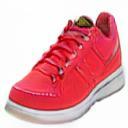} \\

\end{tabular}
\end{center}
\caption{Comparison between our method and StarGAN-v2 on Edges2Shoes. Although both methods perform well on translating edges to shoes, StarGAN-v2 fails to disentangle content from style in the shoes domain and fails to transfer style between shoe images.}  
\label{fig:edges2shoes}
\vspace{-1em}
\end{figure}

Another evidence for the content leakage of StarGAN-v2 is demonstrated in Fig.~\ref{fig:edges2shoes}. In the case of translating edges to shoes, both our method and StarGAN-v2 succeed in translating to the correct style in the target domain. However, in the case of translating between shoe images within the same domain, StarGAN-v2 leaks the entire input image through the content code and can barely change its color. The effectiveness of our proposed content-bottleneck term can be easily observed as our model performs well and is able to translate the style of the shoe while preserving the structure of the image.

Additional results are provided in the supplementary material.

\section{Discussion}

\paragraph{VAEs for disentanglement in image translation models}
Recent image translation methods rely on adversarial objectives for learning a domain-invariant content representation. To the best of our knowledge, a VAE-based content bottleneck has not been used to disentangle content from style in multi-modal image translation before. Exceptions that we are aware of are: UNIT \cite{liu2017unsupervised} and LORD \cite{gabbay2019demystifying}. UNIT does not tackle the multi-modal setting and does not specifically use VAE for disentanglement. LORD specifically uses a content bottleneck for content-class disentanglement in a non-adversarial framework but has not considered the multi-modal setting in which domains exhibit style variations.

\paragraph{Representation Disentanglement}
We have significantly improved the disentanglement in the generated images over a state-of-the-art image translation framework. However, in preliminary experiments we observed that the disentanglement is not well satisfied at the representation level (e.g. the domain can still be classified from the content codes), in our method and the baselines. LORD tackles this issue \cite{gabbay2019demystifying} and utilizes latent optimization for learning disentangled representations. Unfortunately, it can not model intra-domain variations and scale to high resolutions. We leave this challenge to future work.

\paragraph{Integration with other state-of-the-art models}
We have also made an effort to integrate our content-bottleneck principle with other domain translation frameworks such as MUNIT. Although some considerable improvement has been achieved in reducing style-leakage (especially in translating dogs to wildlife), MUNIT suffers from other instability issues and therefore could not compete with our main framework. 

\paragraph{Evaluation metrics for style-content disentanglement}
As stated in the experimental section, current benchmarks for style-content disentanglement are not equipped with well-designed metrics for measuring style disentanglement, primarily due to their unsupervised nature. We believe that the development of better evaluation criteria will greatly speed-up progress in this field.

\section{Conclusion}
We present a simple yet principled approach for improving style-content disentanglement in image translation. We show the effectiveness of our proposed content-bottleneck in preventing style leakage and improving the translation performance. Our method produces state-of-the-art results in terms of visual quality and output style diversity.


\clearpage

\bibliographystyle{plainnat}
\bibliography{neurips_2020}

\begin{thebibliography}{21}
\providecommand{\natexlab}[1]{#1}
\providecommand{\url}[1]{\texttt{#1}}
\expandafter\ifx\csname urlstyle\endcsname\relax
  \providecommand{\doi}[1]{doi: #1}\else
  \providecommand{\doi}{doi: \begingroup \urlstyle{rm}\Url}\fi

\bibitem[Benaim et~al.(2019)Benaim, Khaitov, Galanti, and
  Wolf]{benaim2019domain}
Sagie Benaim, Michael Khaitov, Tomer Galanti, and Lior Wolf.
\newblock Domain intersection and domain difference.
\newblock In \emph{Proceedings of the IEEE International Conference on Computer
  Vision}, pages 3445--3453, 2019.

\bibitem[Bouchacourt et~al.(2018)Bouchacourt, Tomioka, and
  Nowozin]{bouchacourt2018multi}
Diane Bouchacourt, Ryota Tomioka, and Sebastian Nowozin.
\newblock Multi-level variational autoencoder: Learning disentangled
  representations from grouped observations.
\newblock In \emph{Thirty-Second AAAI Conference on Artificial Intelligence},
  2018.

\bibitem[Choi et~al.(2018)Choi, Choi, Kim, Ha, Kim, and Choo]{choi2018stargan}
Yunjey Choi, Minje Choi, Munyoung Kim, Jung-Woo Ha, Sunghun Kim, and Jaegul
  Choo.
\newblock Stargan: Unified generative adversarial networks for multi-domain
  image-to-image translation.
\newblock In \emph{Proceedings of the IEEE conference on computer vision and
  pattern recognition}, pages 8789--8797, 2018.

\bibitem[Choi et~al.(2020)Choi, Uh, Yoo, and Ha]{choi2019stargan}
Yunjey Choi, Youngjung Uh, Jaejun Yoo, and Jung-Woo Ha.
\newblock Stargan v2: Diverse image synthesis for multiple domains.
\newblock In \emph{Proceedings of the IEEE conference on computer vision and
  pattern recognition}, 2020.

\bibitem[Denton et~al.(2017)]{denton2017unsupervised}
Emily~L Denton et~al.
\newblock Unsupervised learning of disentangled representations from video.
\newblock In \emph{Advances in neural information processing systems}, pages
  4414--4423, 2017.

\bibitem[Gabbay and Hoshen(2020)]{gabbay2019demystifying}
Aviv Gabbay and Yedid Hoshen.
\newblock Demystifying inter-class disentanglement.
\newblock In \emph{ICLR}, 2020.

\bibitem[Harsh~Jha et~al.(2018)Harsh~Jha, Anand, Singh, and
  Veeravasarapu]{harsh2018disentangling}
Ananya Harsh~Jha, Saket Anand, Maneesh Singh, and VSR Veeravasarapu.
\newblock Disentangling factors of variation with cycle-consistent variational
  auto-encoders.
\newblock In \emph{ECCV}, 2018.

\bibitem[Heusel et~al.(2017)Heusel, Ramsauer, Unterthiner, Nessler, and
  Hochreiter]{heusel2017fid}
Martin Heusel, Hubert Ramsauer, Thomas Unterthiner, Bernhard Nessler, and Sepp
  Hochreiter.
\newblock Gans trained by a two time-scale update rule converge to a local nash
  equilibrium.
\newblock In \emph{Advances in neural information processing systems}, pages
  6626--6637, 2017.

\bibitem[Huang and Belongie(2017)]{huang2017arbitrary}
Xun Huang and Serge Belongie.
\newblock Arbitrary style transfer in real-time with adaptive instance
  normalization.
\newblock In \emph{Proceedings of the IEEE International Conference on Computer
  Vision}, pages 1501--1510, 2017.

\bibitem[Huang et~al.(2018)Huang, Liu, Belongie, and
  Kautz]{huang2018multimodal}
Xun Huang, Ming-Yu Liu, Serge Belongie, and Jan Kautz.
\newblock Multimodal unsupervised image-to-image translation.
\newblock In \emph{Proceedings of the European Conference on Computer Vision
  (ECCV)}, pages 172--189, 2018.

\bibitem[Isola et~al.(2017)Isola, Zhu, Zhou, and Efros]{isola2017image}
Phillip Isola, Jun-Yan Zhu, Tinghui Zhou, and Alexei~A Efros.
\newblock Image-to-image translation with conditional adversarial networks.
\newblock In \emph{Proceedings of the IEEE conference on computer vision and
  pattern recognition}, pages 1125--1134, 2017.

\bibitem[Karras et~al.(2019)Karras, Laine, and Aila]{karras2019style}
Tero Karras, Samuli Laine, and Timo Aila.
\newblock A style-based generator architecture for generative adversarial
  networks.
\newblock In \emph{Proceedings of the IEEE Conference on Computer Vision and
  Pattern Recognition}, pages 4401--4410, 2019.

\bibitem[Liu et~al.(2017)Liu, Breuel, and Kautz]{liu2017unsupervised}
Ming-Yu Liu, Thomas Breuel, and Jan Kautz.
\newblock Unsupervised image-to-image translation networks.
\newblock In \emph{Advances in neural information processing systems}, pages
  700--708, 2017.

\bibitem[Liu et~al.(2019)Liu, Huang, Mallya, Karras, Aila, Lehtinen, and
  Kautz]{liu2019few}
Ming-Yu Liu, Xun Huang, Arun Mallya, Tero Karras, Timo Aila, Jaakko Lehtinen,
  and Jan Kautz.
\newblock Few-shot unsupervised image-to-image translation.
\newblock In \emph{Proceedings of the IEEE International Conference on Computer
  Vision}, pages 10551--10560, 2019.

\bibitem[Mathieu et~al.(2016)Mathieu, Zhao, Zhao, Ramesh, Sprechmann, and
  LeCun]{mathieu2016disentangling}
Michael~F Mathieu, Junbo~Jake Zhao, Junbo Zhao, Aditya Ramesh, Pablo
  Sprechmann, and Yann LeCun.
\newblock Disentangling factors of variation in deep representation using
  adversarial training.
\newblock In \emph{NIPS}, 2016.

\bibitem[Szab{\'o} et~al.(2018)Szab{\'o}, Hu, Portenier, Zwicker, and
  Favaro]{szabo2017challenges}
Attila Szab{\'o}, Qiyang Hu, Tiziano Portenier, Matthias Zwicker, and Paolo
  Favaro.
\newblock Challenges in disentangling independent factors of variation.
\newblock \emph{ICLRW}, 2018.

\bibitem[Wah et~al.(2011)Wah, Branson, Welinder, Perona, and
  Belongie]{wah2011caltech}
Catherine Wah, Steve Branson, Peter Welinder, Pietro Perona, and Serge
  Belongie.
\newblock The caltech-ucsd birds-200-2011 dataset.
\newblock 2011.

\bibitem[Wang et~al.(2018)Wang, Liu, Zhu, Tao, Kautz, and
  Catanzaro]{wang2018high}
Ting-Chun Wang, Ming-Yu Liu, Jun-Yan Zhu, Andrew Tao, Jan Kautz, and Bryan
  Catanzaro.
\newblock High-resolution image synthesis and semantic manipulation with
  conditional gans.
\newblock In \emph{Proceedings of the IEEE conference on computer vision and
  pattern recognition}, pages 8798--8807, 2018.

\bibitem[Yu and Grauman(2014)]{yu2014fine}
Aron Yu and Kristen Grauman.
\newblock Fine-grained visual comparisons with local learning.
\newblock In \emph{Proceedings of the IEEE Conference on Computer Vision and
  Pattern Recognition}, pages 192--199, 2014.

\bibitem[Zhang et~al.(2018)Zhang, Isola, Efros, Shechtman, and
  Wang]{zhang2018unreasonable}
Richard Zhang, Phillip Isola, Alexei~A Efros, Eli Shechtman, and Oliver Wang.
\newblock The unreasonable effectiveness of deep features as a perceptual
  metric.
\newblock In \emph{Proceedings of the IEEE Conference on Computer Vision and
  Pattern Recognition}, pages 586--595, 2018.

\bibitem[Zhu et~al.(2017)Zhu, Park, Isola, and Efros]{zhu2017unpaired}
Jun-Yan Zhu, Taesung Park, Phillip Isola, and Alexei~A Efros.
\newblock Unpaired image-to-image translation using cycle-consistent
  adversarial networks.
\newblock In \emph{Proceedings of the IEEE international conference on computer
  vision}, pages 2223--2232, 2017.

\end{thebibliography}

\clearpage

\appendix
\section{Appendix}

\subsection{Qualitative results}

\begin{figure}
\begin{center}

\begin{tabular}{@{\hskip0pt}c@{\hskip0pt}c@{\hskip0pt}c@{\hskip0pt}c@{\hskip0pt}c@{\hskip0pt}c}
& ~~~~~~~~~~~~~\textbf{Style} ~~~~~~ \begin{turn}{90} ~~~~~~~ \textbf{Content}\end{turn} ~ & \includegraphics[width=0.19\linewidth]{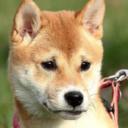} & \includegraphics[width=0.19\linewidth]{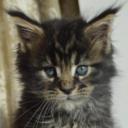} &
\includegraphics[width=0.19\linewidth]{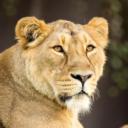} & \includegraphics[width=0.19\linewidth]{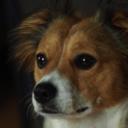} \\

\begin{turn}{90} ~~~~ StarGAN-v2\end{turn} & 
\includegraphics[width=0.19\linewidth]{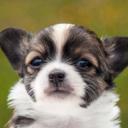} & 
\includegraphics[width=0.19\linewidth]{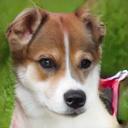} & \includegraphics[width=0.19\linewidth]{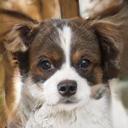} & \includegraphics[width=0.19\linewidth]{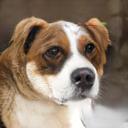} &
\includegraphics[width=0.19\linewidth]{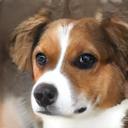} \\
\begin{turn}{90} ~~~~~~~~~ \textbf{Ours}\end{turn} & 
 &
 \includegraphics[width=0.19\linewidth]{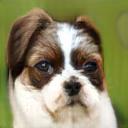} & \includegraphics[width=0.19\linewidth]{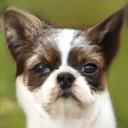} & \includegraphics[width=0.19\linewidth]{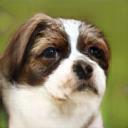} &
\includegraphics[width=0.19\linewidth]{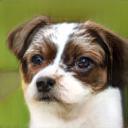} \\

\begin{turn}{90} ~~~~ StarGAN-v2\end{turn} & 
\includegraphics[width=0.19\linewidth]{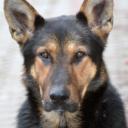} & \includegraphics[width=0.19\linewidth]{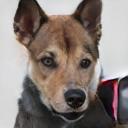} & \includegraphics[width=0.19\linewidth]{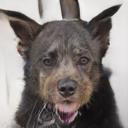} & \includegraphics[width=0.19\linewidth]{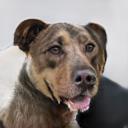} &
\includegraphics[width=0.19\linewidth]{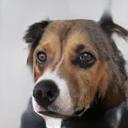} \\
\begin{turn}{90} ~~~~~~~~~ \textbf{Ours}\end{turn} & 
 &
 \includegraphics[width=0.19\linewidth]{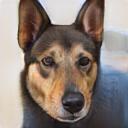} & \includegraphics[width=0.19\linewidth]{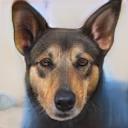} & \includegraphics[width=0.19\linewidth]{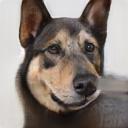} &
\includegraphics[width=0.19\linewidth]{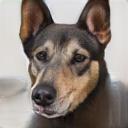} \\

\begin{turn}{90} ~~~~ StarGAN-v2\end{turn} & 
\includegraphics[width=0.19\linewidth]{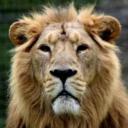} & \includegraphics[width=0.19\linewidth]{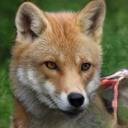} & \includegraphics[width=0.19\linewidth]{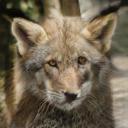} & \includegraphics[width=0.19\linewidth]{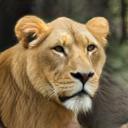} &
\includegraphics[width=0.19\linewidth]{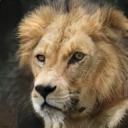} \\
\begin{turn}{90} ~~~~~~~~~ \textbf{Ours}\end{turn} & 
 &
 \includegraphics[width=0.19\linewidth]{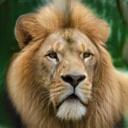} & \includegraphics[width=0.19\linewidth]{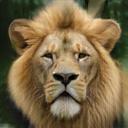} & \includegraphics[width=0.19\linewidth]{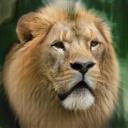} &
\includegraphics[width=0.19\linewidth]{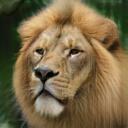} \\

\end{tabular}
\end{center}
\caption{More qualitative results of the comparison between our method and StarGAN-v2 on AFHQ. StarGAN-v2 leaks a significant amount of details of the content image and generates unreliable and inconsistent translations. Our method produces much more disentangled results and captures the target style faithfully.}
\label{fig:afhq_ref_2}
\end{figure}

\begin{figure}
\begin{center}

\begin{tabular}{@{\hskip0pt}c@{\hskip0pt}c@{\hskip0pt}c@{\hskip0pt}c@{\hskip0pt}c@{\hskip0pt}c}
& ~~~~~~~~~~~~~\textbf{Style} ~~~~~~ \begin{turn}{90} ~~~~~~~ \textbf{Content}\end{turn} ~ & \includegraphics[width=0.19\linewidth]{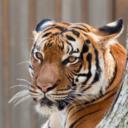} & \includegraphics[width=0.19\linewidth]{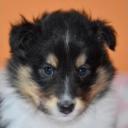} &
\includegraphics[width=0.19\linewidth]{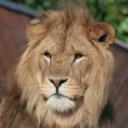} & \includegraphics[width=0.19\linewidth]{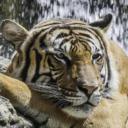} \\

\begin{turn}{90} ~~~~ StarGAN-v2\end{turn} & 
\includegraphics[width=0.19\linewidth]{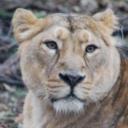} & 
\includegraphics[width=0.19\linewidth]{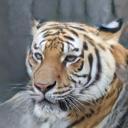} & \includegraphics[width=0.19\linewidth]{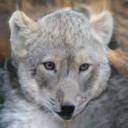} & \includegraphics[width=0.19\linewidth]{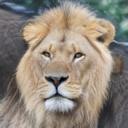} &
\includegraphics[width=0.19\linewidth]{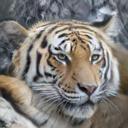} \\
\begin{turn}{90} ~~~~~~~~~ \textbf{Ours}\end{turn} & 
 &
 \includegraphics[width=0.19\linewidth]{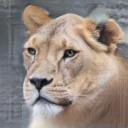} & \includegraphics[width=0.19\linewidth]{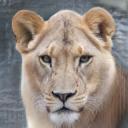} & \includegraphics[width=0.19\linewidth]{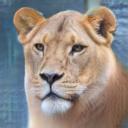} &
\includegraphics[width=0.19\linewidth]{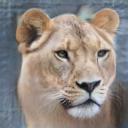} \\

\begin{turn}{90} ~~~~ StarGAN-v2\end{turn} & 
\includegraphics[width=0.19\linewidth]{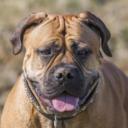} & \includegraphics[width=0.19\linewidth]{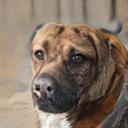} & \includegraphics[width=0.19\linewidth]{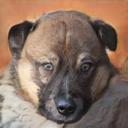} & \includegraphics[width=0.19\linewidth]{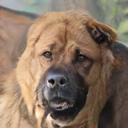} &
\includegraphics[width=0.19\linewidth]{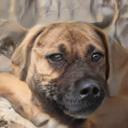} \\
\begin{turn}{90} ~~~~~~~~~ \textbf{Ours}\end{turn} & 
 &
 \includegraphics[width=0.19\linewidth]{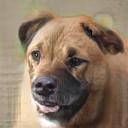} & \includegraphics[width=0.19\linewidth]{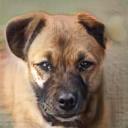} & \includegraphics[width=0.19\linewidth]{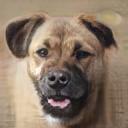} &
\includegraphics[width=0.19\linewidth]{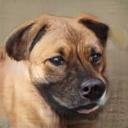} \\

\begin{turn}{90} ~~~~ StarGAN-v2\end{turn} & 
\includegraphics[width=0.19\linewidth]{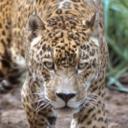} & \includegraphics[width=0.19\linewidth]{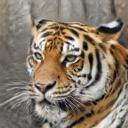} & \includegraphics[width=0.19\linewidth]{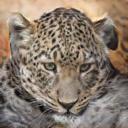} & \includegraphics[width=0.19\linewidth]{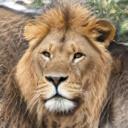} &
\includegraphics[width=0.19\linewidth]{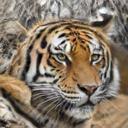} \\
\begin{turn}{90} ~~~~~~~~~ \textbf{Ours}\end{turn} & 
 &
 \includegraphics[width=0.19\linewidth]{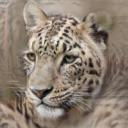} & \includegraphics[width=0.19\linewidth]{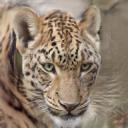} & \includegraphics[width=0.19\linewidth]{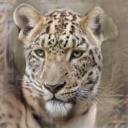} &
\includegraphics[width=0.19\linewidth]{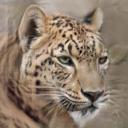} \\

\end{tabular}
\end{center}
\caption{More qualitative results of the comparison between our method and StarGAN-v2 on AFHQ. StarGAN-v2 leaks a significant amount of details of the content image and generates unreliable and inconsistent translations. Our method produces much more disentangled results and captures the target style faithfully.}
\label{fig:afhq_ref_3}
\end{figure}

\begin{figure}
\begin{center}

\begin{tabular}{@{\hskip0pt}c@{\hskip0pt}c@{\hskip0pt}c@{\hskip0pt}c@{\hskip0pt}c@{\hskip0pt}c}
& ~~~~~~~~~~~~~\textbf{Style} ~~~~~~ \begin{turn}{90} ~~~~~~~ \textbf{Content}\end{turn} ~ & \includegraphics[width=0.19\linewidth]{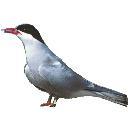} & \includegraphics[width=0.19\linewidth]{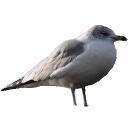} 0&
\includegraphics[width=0.19\linewidth]{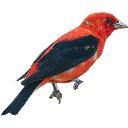} & \includegraphics[width=0.19\linewidth]{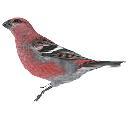} \\

\begin{turn}{90} ~~~~ StarGAN-v2\end{turn} & 
\includegraphics[width=0.19\linewidth]{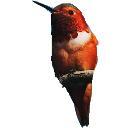} & 
\includegraphics[width=0.19\linewidth]{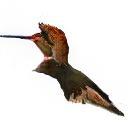} & \includegraphics[width=0.19\linewidth]{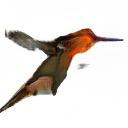} & \includegraphics[width=0.19\linewidth]{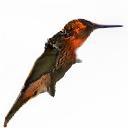} &
\includegraphics[width=0.19\linewidth]{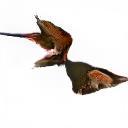} \\
\begin{turn}{90} ~~~~~~~~~ \textbf{Ours}\end{turn} & 
 &
 \includegraphics[width=0.19\linewidth]{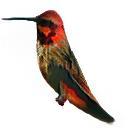} & \includegraphics[width=0.19\linewidth]{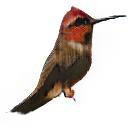} & \includegraphics[width=0.19\linewidth]{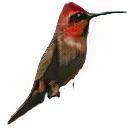} &
\includegraphics[width=0.19\linewidth]{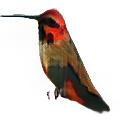} \\

\begin{turn}{90} ~~~~ StarGAN-v2\end{turn} & 
\includegraphics[width=0.19\linewidth]{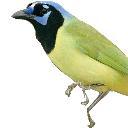} & 
\includegraphics[width=0.19\linewidth]{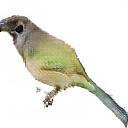} & \includegraphics[width=0.19\linewidth]{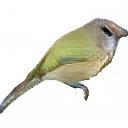} & \includegraphics[width=0.19\linewidth]{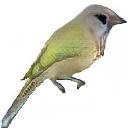} &
\includegraphics[width=0.19\linewidth]{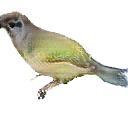} \\
\begin{turn}{90} ~~~~~~~~~ \textbf{Ours}\end{turn} & 
 &
 \includegraphics[width=0.19\linewidth]{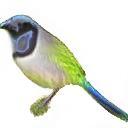} & \includegraphics[width=0.19\linewidth]{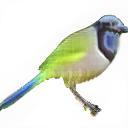} & \includegraphics[width=0.19\linewidth]{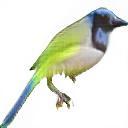} &
\includegraphics[width=0.19\linewidth]{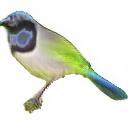} \\

\begin{turn}{90} ~~~~ StarGAN-v2\end{turn} & 
\includegraphics[width=0.19\linewidth]{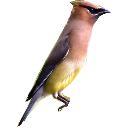} & 
\includegraphics[width=0.19\linewidth]{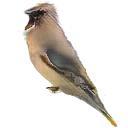} & \includegraphics[width=0.19\linewidth]{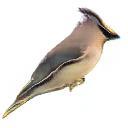} & \includegraphics[width=0.19\linewidth]{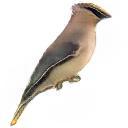} &
\includegraphics[width=0.19\linewidth]{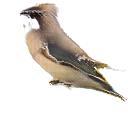} \\
\begin{turn}{90} ~~~~~~~~~ \textbf{Ours}\end{turn} & 
 &
 \includegraphics[width=0.19\linewidth]{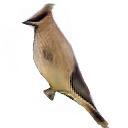} & \includegraphics[width=0.19\linewidth]{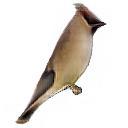} & \includegraphics[width=0.19\linewidth]{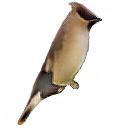} &
\includegraphics[width=0.19\linewidth]{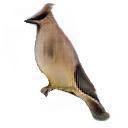} \\

\end{tabular}
\end{center}
\caption{More qualitative results of the comparison between our method and StarGAN-v2 on CUB-47. It can be seen that StarGAN-v2 tends to overfit the content exactly and often fails to generate valid bird images of the target style. Moreover, our method better captures the fine-grained details of the target style. For example, it is able to transfer-over the throat color and the head pattern.}
\label{fig:cub_ref_2}
\end{figure}

\begin{figure}[t]
\begin{center}

\begin{tabular}{@{\hskip0pt}c@{\hskip0pt}c@{\hskip0pt}c@{\hskip0pt}c@{\hskip0pt}c@{\hskip0pt}c}
& ~~~~~~~~~~\textbf{Content} ~~~~ \begin{turn}{90} ~~~~~~~ \textbf{Style}\end{turn} ~ & \includegraphics[width=0.19\linewidth]{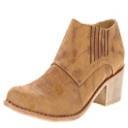} &
\includegraphics[width=0.19\linewidth]{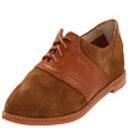} & \includegraphics[width=0.19\linewidth]{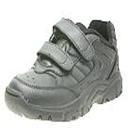} &
\includegraphics[width=0.19\linewidth]{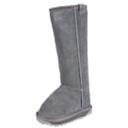} \\

\begin{turn}{90} ~~ StarGAN-v2\end{turn} & 
\includegraphics[width=0.19\linewidth]{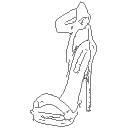} & \includegraphics[width=0.19\linewidth]{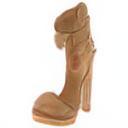} & \includegraphics[width=0.19\linewidth]{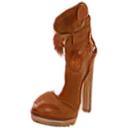} & \includegraphics[width=0.19\linewidth]{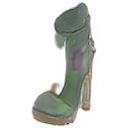} &
\includegraphics[width=0.19\linewidth]{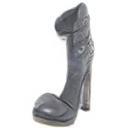} \\
\begin{turn}{90} ~~~~~~ \textbf{Ours}\end{turn} & 
 &
 \includegraphics[width=0.19\linewidth]{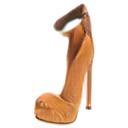} & \includegraphics[width=0.19\linewidth]{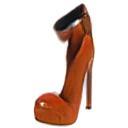} & \includegraphics[width=0.19\linewidth]{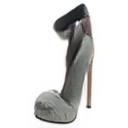} &
\includegraphics[width=0.19\linewidth]{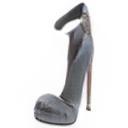} \\

\begin{turn}{90} ~~ StarGAN-v2\end{turn} & 
\includegraphics[width=0.19\linewidth]{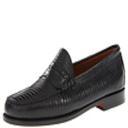} & \includegraphics[width=0.19\linewidth]{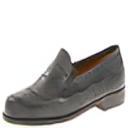} & \includegraphics[width=0.19\linewidth]{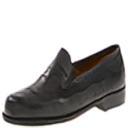} & \includegraphics[width=0.19\linewidth]{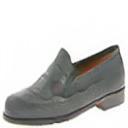} &
\includegraphics[width=0.19\linewidth]{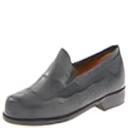} \\
\begin{turn}{90} ~~~~~~ \textbf{Ours}\end{turn} & 
 &
 \includegraphics[width=0.19\linewidth]{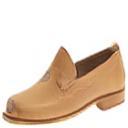} & \includegraphics[width=0.19\linewidth]{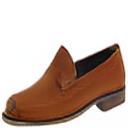} & \includegraphics[width=0.19\linewidth]{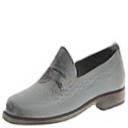} &
\includegraphics[width=0.19\linewidth]{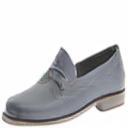} \\

\begin{turn}{90} ~~ StarGAN-v2\end{turn} & 
\includegraphics[width=0.19\linewidth]{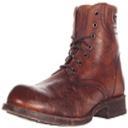} & \includegraphics[width=0.19\linewidth]{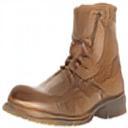} & \includegraphics[width=0.19\linewidth]{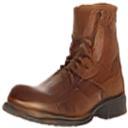} & \includegraphics[width=0.19\linewidth]{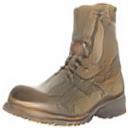} &
\includegraphics[width=0.19\linewidth]{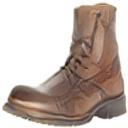} \\
\begin{turn}{90} ~~~~~~ \textbf{Ours}\end{turn} & 
 &
 \includegraphics[width=0.19\linewidth]{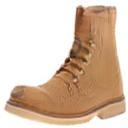} & \includegraphics[width=0.19\linewidth]{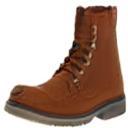} & \includegraphics[width=0.19\linewidth]{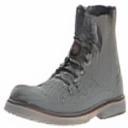} &
\includegraphics[width=0.19\linewidth]{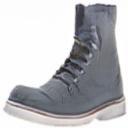} \\

\end{tabular}
\end{center}
\caption{More qualitative results from the comparison between our method and StarGAN-v2 on Edges2Shoes. Although both methods perform well on translating edges to shoes, StarGAN-v2 fails to disentangle content from style in the shoes domain and fails to transfer style between shoe images.}  
\label{fig:edges2shoes_ref2}
\end{figure}

\paragraph{AFHQ}
We provide more qualitative results on AFHQ in Fig.~\ref{fig:afhq_ref_2} and \ref{fig:afhq_ref_3}. It can be seen that especially in translating dogs to wildlife and vice versa, StarGAN-v2 leaks a significant amount of details of the content image and generates unreliable and inconsistent translations. Our method produces much more disentangled results and captures the target style faithfully.

\paragraph{CUB-47}
We provide more qualitative results on CUB-47 in Fig.~\ref{fig:cub_ref_2}.
It can be seen that our method better captures the fine-grained details of the target style. For example, it is able to transfer-over the throat color and the head pattern.

\paragraph{Edges2Shoes}
More evidence for the effectiveness of our proposed content-bottleneck is demonstrated in Fig.~\ref{fig:edges2shoes_ref2}. StarGAN-v2 succeeds to translate edges to shoes as there is no style-related information to leak from the edge images to the shoe images, but fails to disentangle content from style within the shoes domain and barely changes the shoe color. Our method is capable of translating images within a domain and across domains.

\end{document}